\documentclass[sn-mathphys]{sn-jnl}
\jyear{2022}%
\usepackage[utf8]{inputenc}
\usepackage{hyperref}

\begin{document}

\title[Vehicle management in a modular production context using Deep Q-Learning]{Vehicle management in a modular production context using Deep Q-Learning}

\author[1]{\fnm{Lucain} \sur{Pouget}}\email{lucain.pouget@skalarsystems.com}
\equalcont{These authors contributed equally to this work.}

\author[2]{\fnm{Timo} \sur{Hasenbichler}}\email{timo.hasenbichler@protonmail.com}
\equalcont{These authors contributed equally to this work.}

\author[1]{\fnm{Jakob} \sur{Auer}}\email{jakob.auer@skalarsystems.com}

\author[2]{\fnm{Klaus} \sur{Lichtenegger}}\email{klaus.lichtenegger@fh-joanneum.at}

\author[2,3,4,5,6]{\fnm{Andreas} \sur{Windisch}}\email{awindisch@know-center.at}

\affil*[1]{\orgdiv{Skalar Systems GmbH}, \orgaddress{\street{Dr.-Robert-Graf-Stra\ss e 9}, \city{Graz}, \postcode{8010}, \country{Austria}}}

\affil[2]{\orgdiv{FH JOANNEUM -- University of Applied Sciences}, \orgname{Data Science and Artificial Intelligence}, \orgaddress{\street{Eckertstra\ss e 30i}, \city{Graz}, \postcode{8020}, \country{Austria}}}

\affil[3]{\orgdiv{Know-Center GmbH}, \orgaddress{\street{Inffeldgasse 13}, \city{Graz}, \postcode{8010}, \country{Austria}}}

\affil[4]{\orgdiv{Institute of Interactive Systems and Data Science}, \orgname{Graz University of Technology}, \orgaddress{\street{Inffeldgasse 13}, \city{Graz}, \postcode{8010}, \country{Austria}}}

\affil[5]{\orgdiv{Physics Department}, \orgname{Washington University in St. Louis}, \orgaddress{\street{One Brookings Drive}, \state{MO}, \city{St. Louis}, \postcode{63130}, \country{USA}}}

\affil[6]{\orgdiv{RL Community}, \orgname{AI Austria}, \orgaddress{\street{Wollzeile 24/12}, \city{Vienna}, \postcode{1010}, \country{Austria}}}

\abstract{We investigate the feasibility of deploying Deep-Q based deep reinforcement learning agents to job-shop scheduling problems in the context of modular production facilities, using discrete event simulations for the environment. These environments are comprised of a source and sink for the parts to be processed, as well as (several) workstations. The agents are trained to schedule automated guided vehicles to transport the parts back and forth between those stations in an optimal fashion. Starting from a very simplistic setup, we increase the complexity of the environment and compare the agents' performances with well established heuristic approaches, such as first-in-first-out based agents, cost tables and a nearest-neighbor approach.

We furthermore seek particular configurations of the environments in which the heuristic approaches struggle, to investigate to what degree the Deep-Q agents are affected by these challenges. We find that Deep-Q based agents show comparable performance as the heuristic baselines. Furthermore, our findings suggest that the DRL agents exhibit an increased robustness to noise, as compared to the conventional approaches. Overall, we find that DRL agents constitute a valuable approach for this type of scheduling problems.}



\maketitle

\section{Introduction}
\label{sec:introduction}

\subsection{Modular Production}
Modular production systems based on Automated Guided Vehicles (AGVs)  allow a drastically increased flexibility of the production plant or setup. Instead of a long line of machines connected by conveyor belts, production islands (consisting of one or several machines) are spread over the production plant and connected via AGVs. The inputs and outputs of these production islands usually also contain buffers.
The increased flexibility allows advancements like multiple product variants in the same plant, alternative stations (two stations of the same type as performance improvement or for failure safety) or the reuse of expensive stations at different stages of the production.

\subsection{The Job Shop Scheduling Problem}
\label{ssec:JSSP}

In modular production environments, the optimized scheduling of the AGVs (the decision which AGV is assigned to which transport task) is crucial for the system's overall performance. This task leads to the Job Shop Scheduling Problem (JSSP), which is -- like the related Travelling Salesman Problem -- a computationally ``hard'' (NP-complete) problem of enormous practical importance, \cite{Garey1976}.
Due to the typically factorial growth of possible arrangement with the size of the problem, exact solutions can only be found for very small problems. Since the AGV to Job scheduling has to be done in real-time, heuristic strategies are used. This is usually done with algorithms like nearest neighbor \cite{10.1007/BFb0032050} or cost tables \cite{https://doi.org/10.1002/nav.3800020109}. All of these methods have various advantages and drawbacks, but they are all reaching a limit with increasing complexity and system size. Varying driving times (because of human machine interaction or other vehicles in the systems) or processing times highly affect the systems performance and stability. Wrong or suboptimal scheduling decisions can lead to deadlocks in the system.
We show that Deep-Q based deep reinforcement  learning agents are capable to overcome the aforementioned problems and also allow the calculation of optimal solutions in real-time.

\subsection{Deep Q-Learning}
\label{ssec:deep_Q}

Reinforcement learning aims to train an agent to interact with the environment in a way that maximizes the expected cumulative reward. The policy of the agent  $\pi(\theta)$ is defined as a function that maps from a given state $s$ to a suitable action $a$.
Q-learning is a value-based approach, which means that the agent will try to approximate the expected cumulative reward (the “value”) for each possible pair of (state/action). This value is formalized as
\begin{equation}
Q^{*}(s, a)=\mathbb{E}_{s^{\prime}}\left[r+\gamma \max _{a^{\prime}} Q^{*}\left(s^{\prime}, a^{\prime}\right) \mid s, a\right]
\end{equation}
The expected cumulative reward $Q(s,\,a)$ in state $s$ for an action $a$ is defined as the sum of the immediate reward $r(s,\,a)$ and the highest Q-value for the next states among all possible actions. A discount factor $\gamma\in[0,\,1]$ is applied to the future reward in order to control how to rate future rewards compared to the immediate ones.

Due to the usually large number of state-action pairs, filling the whole table of Q values is possible only for very small problems. An approach to overcome this obstacle is \emph{Deep Q-learning}. This is a type of Q-learning where the value function is approximated by a deep neural network, that is, by a network with several layers. This is justified by the universal function approximation property of sufficiently deep neural networks, \cite{Hornik1989}.

Each time the Agent makes a decision, an experience (initial state, action, reward, next state) is saved into a buffer (the Replay Memory). The weights of the network are trained using backpropagation on mini-batches, sampled from the Replay Memory, \cite{Sutton2018}. 

\subsection{Structure of the article}
\label{ssec:structure}

This paper is structured as follows: In section \ref{sec:environment}, the environment used for training the DRL agents is discussed in detail, including the particular configurations of the environments, as well as technical aspects, such as using OpenAI's gym interface, benchmark settings and heuristic approaches that have been used for comparison. The design of the DRL agent is then discussed in section \ref{sec:design_DRLagent}, where, based on Sec.~\ref{ssec:deep_Q}, a more detailed explanation of the reward signal used for training is given. The main findings of this paper are summarized in section \ref{sec:results}, where all chosen production setups are presented, together with the respective performance of the DRL agents and heuristic approaches. Finally, in section \ref{sec:conclusions}, we conclude with a short overall summary and point out possible paths for future investigations.

\section{Construction of the environment}
\label{sec:environment}

\subsection{Production plant modeling}

A production plant is modelled by a set of $n_{M}$ modular workstations. Each workstation consists of an Input Buffer ($IB$), a Production Unit ($PU$) and an Output Buffer ($OF$). Both buffers have a capacity of $n_{\rm buf}$ and are running as FIFO queues (First In-First Out). In addition, we defined a Start of Line station ($Source$) and an End of Line station ($Sink$). The $Source$ can be seen as a station with only an $OB$ and the $Sink$ as a station with only an $IB$.

Parts are defined by their part type PT. For each part type $\text{PT}_j$, a sequence of $n_{O}$ operations is defined. This sequence represents the workstations that the parts of type $\text{PT}_j$ must visit in the right order. The sequence always starts at the Source and ends at the Sink. The same workstation can be used multiple times in a sequence. Different part types do not necessarily use the same workstations. Parts are carried by $n_V$ AGVs $V$ with speed $v_V$. A vehicle can only carry 1 part at a time. The transfer of a part between a vehicle and a station, or between 2 units of a station, has a duration $T_{\rm transfer}$.

A weighted multi-directional graph $G$, called “waypoint graph", is defined to represent the distances between the stations. The nodes are the waypoints in the production plant through which the vehicles can pass. Each node $N_{j}$ is described by Cartesian coordinates {$x_{j}$, $y_{j}$}. The edges are the paths connecting these locations. Each edge is either unidirectional or bidirectional and has a weight corresponding to the euclidean distance between its nodes. Each buffer of each workstation is linked to a node of the waypoint graph. A node can only be assigned to a maximum of one buffer.

In our model, the release order of parts at the source is modelled by a uniform distribution over the part types. A source clock $C_{\rm source}$ defines when a new part is released to the system. Our goal is to maximize the production throughput by assigning jobs to the AGVs. A job consists of driving to a station $S_a$, picking a part $P_j$, driving to another station $S_b$, dropping the part $P_j$. Jobs must be assigned in such a way as to ensure that there are no deadlocks. This problem is defined as the Vehicle Management problem (VM), which is solved with deep learning (more precisely with proximal policy optimization) in \cite{Mayer2021}.

\subsection{Implementation details}

\subsubsection{Discrete-event simulation}

The modular production plant is simulated in Python using the Simpy library \cite{matloff2008introduction}, a framework to build Discrete-Event Simulations (DES) \cite{osti_6893405}. In a DES, events are triggered by the environment, each of them at a predefined instant in time. Between two consecutive events, the environment is static. We chose this approach because of its efficiency as compared to continuous simulations. Having a simulation running in Python has also been beneficial for the communication between the Agent and the Environment.

We performed a benchmark of our simulation tool in various cases to estimate the speed-up factor of our approach over simulations not relying on discrete event simulation. We defined the complexity of a configuration as the number of AGVs $N_{\rm agvs}$ times the number of Machines $N_{\rm machines}$. Each production plant configuration has been simulated for 12 hours using the FIFO agent. Since our simulation is not computing each intermediate steps when an AGV moves, but instantly jumps to the next relevant event (e.g. AGV reaches destination), the speed-up factor shows a strong dependence on the parameters of the environment, such as distances, AGV speed and processing times. For typical environments used in training, it takes between 360~ms and 2.6~s to simulate 1h of production, depending on the complexity of the environment. The calculation for the benchmark has been conducted on a laptop computer using a single thread (Dell XPS 15, Intel core i7 2.60Hz). Lacking an obvious definition of complexity of the environment renders quantitative speed-up factors somewhat arbitrary. We thus refrain from providing quantitative data at this point, but emphasize, that our DES based approach shows a very clear advantage in simulation time over discretized time simulations, which is a very desirable property simulation environments used for Deep Reinforcement Learning should possess.  

\subsubsection{OpenAI Gym interface}

We use OpenAI’s library Gym \cite{brockman2016openai} to implement the interface between the Simpy simulation and the decision making process. In particular, the library provides utilities to handle observations, actions and rewards. The main concepts handled by the library are:

\begin{itemize}
    \item \textbf{Observation:} object that describes the state of the Simpy simulation at a given instant. In our case, the Observation is a human-readable dictionary of values. Gym library takes care of serializing this data into a vector that can be sent to the neural network.
    \item \textbf{Action:} object that describes the action taken by an Agent.
    \item \textbf{Reward:} float value describing the amount of reward for a given action. The goal of the training is always to maximize the expected, cumulative reward.
    \item \textbf{Environment:} the minimal acceptable environment is an object implementing a “step” method. This “step” method takes as input an action and outputs a tuple of four elements:
    \begin{itemize}
      \item \textbf{Observation:} the state of the environment once the action has been performed.
      \item \textbf{Reward:} amount of reward achieved by the previous action
      \item \textbf{Done:} a boolean value set to true if the environment must be reset.
      \item \textbf{Info:} any information useful for debugging. Not used in our case.
    \end{itemize}
\end{itemize}

Each time the "step" method of the environment is called, a new Experience -- a tuple (initial state, action taken, next state, reward, done) -- is stored in the memory for the learning process of the agent.

We had two major benefits of using this framework:
\begin{itemize}
    \item Gym contains utilities in order to serialize and deserialize the Observation and Action concepts from a human-readable object to a vector suitable for a neural network.
    \item Gym implements several well known problems \cite{openaigymenvs}. These tasks proved useful for testing and debugging the implementation of our agent against environments already solved by Deep-Q agents.
\end{itemize}

\subsubsection{Communication protocol}

The “Agent” is the algorithm that makes the decisions. We defined a common interface for both, deterministic and parameterized agents, which allowed us to seamlessly plug them into the simulation for comparison. Deterministic agents are algorithms that are entirely hard-coded, while parameterized agents require a training phase before evaluation.

The Agent Interface uses two methods:
\begin{itemize}
    \item \textbf{Act:} takes as input an Observation state describing the environment and returns the action to perform.
    \item \textbf{Step:} takes as input an Experience and returns nothing. This method is used for training the parameterized agents in order to improve their policy.
\end{itemize}

The Controller is an element of the Simpy environment that makes the connection between the AGVs and the Agent. In a production environment, this would be the Fleet Management System.

The communication process is the following:

\begin{enumerate}
    \item The environment is initialized with all its elements.
    \item The simulation starts.
    \item When an AGV is available (it has finished its current task), it triggers a custom Simpy Event. 
    \item The Controller stops the simulation and computes an Observation state from the environment.
    \item The Observation state is sent to the Agent.
    \item The Agent takes an action that is communicated to the Controller.
    \item The Controller deserializes the action and assigns the corresponding job to the waiting AGV. A job is a station from which an AGV must pick up a part.
    \begin{enumerate}
        \item If multiple AGVs are waiting at the same time, the Controller sends the job to the waiting AGV that will perform the action the fastest (i.e. the closest AGV for that job).
        \item The Agent is then called immediately after to define another job for the other waiting AGVs until all AGVs have a job.
    \end{enumerate}
    \item Once all the AGVs have a job, the simulation starts over at 2.
\end{enumerate}

The Agent also has the possibility to choose not to assign any task. In that case, the waiting AGV will stay inactive until a new event is triggered.

\subsection{Benchmark settings}

\subsubsection{Production plant settings}

Several production plants have been configured in order to compare the Agents on different types of problems.

\paragraph{Mayer-Classen-Endisch}

This configuration is defined in \cite{Mayer2021}. It has 2 machines, 1 AGV and 1 part type. We re-implemented the simulation, taking the same parameters to verify that our approach reproduces the results of \cite{Mayer2021}.

\paragraph{1-machine-big}

\begin{figure}[ht]
  \includegraphics[width=1.0\textwidth]{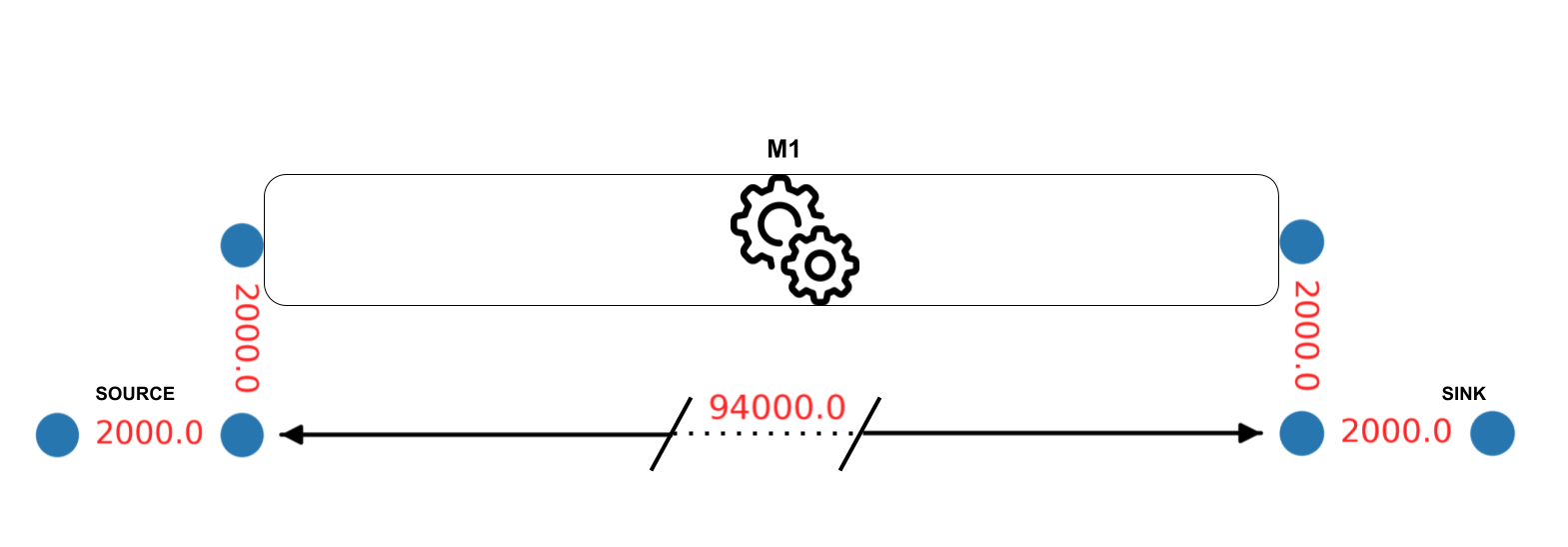}
  \caption{1-machine-big configuration: optimizing the driving time is essential here.}
  \label{fig:plan_1_machine_big}
\end{figure}

This environment has 1 machine, 2 AGVs and 1 part type. The particularity lies in the distances between the nodes of the waypoint graph (see figure \ref{fig:plan_1_machine_big}). The pairs “source/machines input” and “machines output/sink” are very close, while the pairs “source/sink” and “machines input/machines output” are very far apart. Since there are 2 AGVs, the optimal solution is to have one vehicle doing circles between the source and the machine's input, while another vehicle does the same between the machine's output and the sink. We introduced this environment, expecting that the FIFO Agent would struggle with this setup.

\paragraph{3-machines-loop}

\begin{figure}[ht]
  \includegraphics[width=1.0\textwidth]{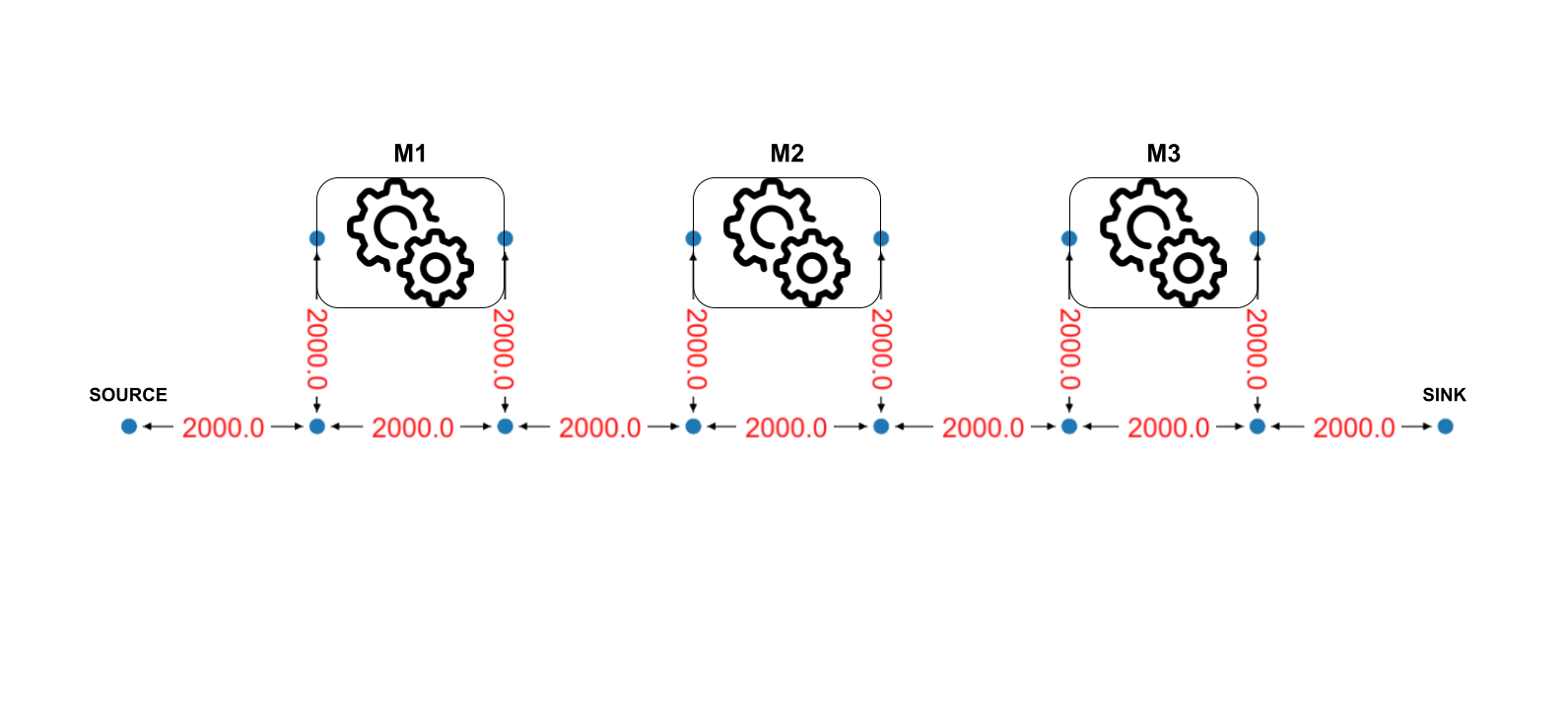}
  \caption{3 machines configuration: the sequence of operations in this setup includes a loop.}
  \label{fig:plan_3_machines}
\end{figure}

We set up a configuration with 3 machines and a unique part type (figure \ref{fig:plan_3_machines}). The particularity here is, that the sequence of operations to perform on a part has a loop, passing twice through the same machine. Each part $P_{j}$ has to follow the sequence $\text{Source} \rightarrow M1 \rightarrow M2 \rightarrow M1 \rightarrow M3 \rightarrow \text{Sink}$. This is a well known type of problem in modular production that often leads to deadlocks when using static algorithms.

\paragraph{6-machines-grid}

\begin{figure}[ht]
  \includegraphics[width=1.0\textwidth]{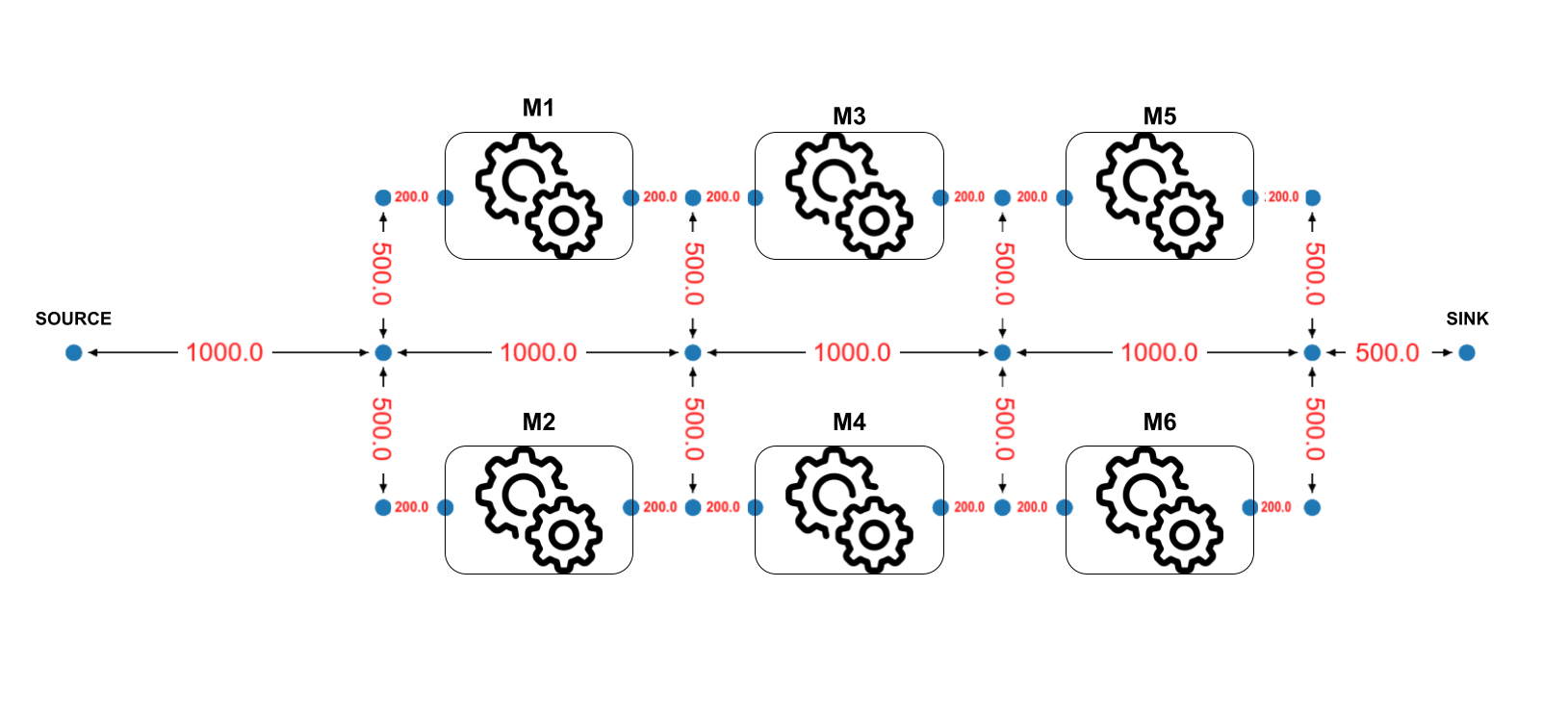}
  \caption{6-machines grid configuration: depending on the scenario, not all machines are in use.}
  \label{fig:plan_grid_6_machines}
\end{figure}

This is a family of problems starting with the same production plant configuration. The goal is to incrementally increase the difficulty of the scenarios and monitor how it impacts the performance of the agent at each step. For all sub-problems, the production plant layout, shown in figure \ref{fig:plan_grid_6_machines}, is shared. It has 6 machines distributed as a grid (2 rows of 3 machines), but depending on the scenario, not all of them are in use.

Complexity is increased in 2 ways:
\begin{itemize}
    \item By increasing the number of machines to visit. Each sub-problem has 2 part types that follow similar paths in the production plant. 3 scenarios are considered, using either 2 machines ($M1 \rightarrow M2$ and $M2 \rightarrow M1$), 4 machines ($M1 \rightarrow M3 \rightarrow M2 \rightarrow M4$ and $M2 \rightarrow M4 \rightarrow M1 \rightarrow M3$) or 6 machines ($M1 \rightarrow M3 \rightarrow M5 \rightarrow M2 \rightarrow M4 \rightarrow M6$ and $M2 \rightarrow M4 \rightarrow M6 \rightarrow M1 \rightarrow M3 \rightarrow M5$).
    \item By increasing the number of AGVs (from 1 to 4). With only 1 vehicle, the limiting factor is usually the speed of the vehicle, whereas with a larger number of vehicles it is possible to reach a 100\% workload on the machines.
\end{itemize}

\subsubsection{Deterministic Agents}

Deterministic agents follow a hard-coded algorithm and are used as a baseline for result comparison. In this section, we call $\text{AGV}_{\rm longest}$ the AGV that has been waiting the longest for a new assignment and $P_{\rm longest}$ the part that has been waiting the longest in the output buffer of a station (named $S_{\rm longest}$).

An AGV is considered as active, if it has a task assigned to it and inactive otherwise.

\paragraph{FIFO}

The First-In First-Out (FIFO) algorithm consists of assigning $S_{\rm longest}$ to $\text{AGV}_{\rm longest}$.

\paragraph{Nearest Neighbor}

The Nearest Neighbor (NN) algorithm consists of selecting $S_{\rm longest}$ and compute for all the AGVs the time it will take to reach the station exit containing this part:
\begin{itemize}
    \item If an AGV is inactive, it estimates the time it will take to drive to $S_{\rm longest}$.
    \item If an AGV is active, it estimates the time it will take to finish the current task and then reach $S_{\rm longest}$.
\end{itemize}

The Agent assigns the job to the closest AGV in terms of duration. If this AGV is already active, the Agent doesn’t assign any task.

\paragraph{Cost Table}

The Cost Table algorithm is an extension of the Nearest Neighbor approach. Instead of doing an estimation only for the station $S_{\rm longest}$, it builds an array of size \texttt{[Number of inactive AGVs, Number of waiting stations]}, containing the time estimation for every AGV/Station pair (i.e., “costs”). The Agent solves the linear sum assignment problem on this Cost Table using the linear sum assignment algorithm \cite{linearsumassignment} implemented in SciPy \cite{2020SciPy-NMeth}. The solution of the problem is a list of AGV/Station pairs that minimizes the duration to start all the tasks. Since the Agent can only make one assignment at a time, it arbitrarily selects the task with the AGV that has been waiting the longest among the AGVs of the solution. The Agent is immediately called again to re-compute the Cost Table and re-solve the optimization problem until all AGVs are active, or no parts are waiting in a station output.

\subsubsection{Methodology}

Our goal is to be able to compare a trained Agent against static algorithms in order to prove the ability of DRL to tackle the Vehicle Management problem in a modular production environment.

We found out that static algorithms are highly vulnerable to the input clock $C_{\rm source}$. If $C_{\rm source}$ is too low, the AGVs will almost immediately overload the first machine with parts and run into a deadlock. On the contrary, if $C_{\rm source}$ is too high, the optimization problem doesn’t make sense anymore because it is too easy to get an output clock $C_{\rm sink}$ equal to $C_{\rm source}$. 

In order to still be able to have baseline results to compare with, we first find the optimal $C_{\rm source}^*$ for each agent and each environment configuration. $C_{\rm source}^*$ is defined as the lowest input clock for which the simulation doesn’t run into a deadlock situation. We use a dichotomic search algorithm to find this value, rounded to 1 second. By construction, $C_{\rm sink}$ is necessarily equal or greater than $C_{\rm source}^*$. Once $C_{\rm source}^*$ is found, we run the simulation for 12 hours of production, gathering the metrics throuhout the run.

For the DRL Agent, $C_{\rm source}$ is set to $0$ during both, train and test phases. It is assumed that it should learn how to avoid deadlocks by not overloading the first machine.

\section{Design of the DRL Agent}
\label{sec:design_DRLagent}

In this section, we specify the deep-Q learning approach, briefly discussed in Sec.~\ref{ssec:deep_Q}, to our specific situation and discuss some details of its implementation.

\subsection{Observation, action, and reward design}
\label{ssec:observation_action_reward}

\subsubsection{Observation state}

The observation state is the vector describing the Environment that is sent to the Agent. It contains all the information needed to choose the next action. We defined a state $S_v$ for each AGV and a state $S_{\rm unit}$ for each station unit. The environment state $S$ is the concatenation of all the $S_v$ and $S_{\rm unit}$ states.

For each vehicle $V_i$, the state $\text{Sv}_i$ consist of:
\begin{itemize}
    \item \texttt{Action\_state}: current action performed by the vehicle as a 1-hot encoded vector: \texttt{DRIVING}, \texttt{TRANSFERRING\_IN}, \texttt{TRANSFERRING\_OUT}, \texttt{WAITING\_FOR\_ORDER}, \texttt{WAITING\_TO\_DROPDOWN}, \texttt{WAITING\_TO\_PICKUP}
    \item \texttt{Carried part}: a representation of the part that is carried, if any (see below).
    \item \texttt{Current order target}: 1-hot encoded vector representing the destination of $V_i$, if driving. If $V_i$ is not currently driving, it is a zero vector of appropriate dimension.
    \item \texttt{Last visited node}: 1-hot encoded vector representing the last node which $V_i$ has visited.
\end{itemize}

\noindent For each unit $U_i$, the state $\text{Sunit}_i$ consists of:
\begin{itemize}
    \item \texttt{Action state}: current action performed by the unit as a 1-hot encoded vector: \texttt{PROCESSING}, \texttt{TRANSFERRING\_IN}, \texttt{TRANSFERRING\_OUT}, \texttt{WAITING\_TO\_DROPDOWN}, \texttt{WAITING\_TO\_PICKUP}
    \item \texttt{Carried parts}: a representation of each part that is in the unit (see below).
\end{itemize}

For each part is the production plant, either in a station unit or an AGV, a state representation is defined as consisting of:
\begin{itemize}
    \item \texttt{Part\_type}: 1-hot encoded vector representing the part type. The information is useful to determine what are the next operations to be performed on the parts.
    \item \texttt{Part\_completion}: float value between 0 and 1. When a part is at the Source, the completion value is 0 while at the Sink it is set to 1. In between, the completion value is increased linearly each time an action is performed on the part (driving or processing).
    \item \texttt{Part\_next\_station}: 1-hot encoded vector representing the next station the part needs to visit (or Sink if processing is done). In principle, 
    if only \texttt{part\_type} and \texttt{part\_completion} is provided, the Agent should be able to learn the next steps for a part by itself. However, we discovered by experience that providing redundant information helped to increase the overall performance.
\end{itemize}

\subsubsection{Action}

The set of all possible actions is equivalent to the set of stations where it is possible to pick-up a part (i.e., the Source and all the workstations). The Sink is not included in the set of actions, since it’s only possible to drop off a part there. In contrast to \cite{Mayer2021}, the set of all dropdown actions \texttt{A\_drop} is not defined in our case, since the dropdown station is always implicit and only depends on the carried part. This is a limitation in the case of a production plant where the same operation can be performed by 2 different workstations, but it is beyond the scope of this paper.

In addition to this set of actions, we also allow the Agent to take a “do no\-thing” action, for which no task is assigned. The inactive vehicles stay inactive until another event triggers a request to the Agent.

\subsubsection{Reward}

The reward function has been one of the most difficult components to configure in this setup. We realized that the performance of the Agent and its capability to learn are highly dependent on the reward signal. In the end, we built a score function for each environment state $S_t$. From a state $S_t$ and for an action $A_t$ that resulted in the state $S_{t+1}$, we defined the reward $R(S_t, A_t)$ as $\text{Score}(S_{t+1}) - \text{Score}(S_t)$.

We built the score of a state as the sum of 4 components:
\begin{itemize}
    \item Per-part reward $S_{\rm pp}$: a positive reward for each completed part that reaches the Sink.
    \item Per-part completion reward $S_{\rm pp\%}$: a positive reward each time a part is being processed. We define the completion of a part (between 0 and 1) as the percentage of processing steps that have been completed over all steps. When the part is at the Source, the completion is 0, while it is 1 at the Sink. This reward helps the Agent learn about the intermediate steps by providing shorter-term feedback.
    \item Per-assigned decision reward $S_{\rm decisions}$: a positive reward for each decision taken by the Agent that is actually assigned to an AGV. In particular, the “do-nothing” action is not rewarded. We noticed a slight increase in performance when forcing the Agent into being less lazy.
    \item Per-second reward $S_{\rm time}$: a positive reward for each second that has been simulated without encountering a deadlock. 
\end{itemize}

We denote by $N_{\rm decisions}(t)$ the number of assigned decisions since the beginning of the simulation, by $N_{\rm pp}(t)$ the total number of completed parts, and by $N_{\rm pp\%}(t)$ the sum of each part completion. In practice $N_{\rm pp\%}(t) \ge N_{\rm pp}(t)$ at all time.

A difficulty we had has been to balance the 4 components of the score function. Depending on the environment configuration, the expected number of assigned decisions and parts processed were highly variable, whereas the number of simulated seconds was always the same (12-hours when no deadlocks). To balance those differences, we defined 3 configurable values: $E_{\rm pp}$ (expected processed parts), $E_{\rm decisions}$ (expected assigned decisions) and $E_{\rm seconds}$ (expected duration). The final score function is defined as a weighted average of the 4 components:

\begin{equation}
S(t) =\begin{cases} 
K \, \left[ \frac{N_{\rm pp}(t) + N_{pp\%}(t)}{E_{\rm pp}} + \frac{N_{\rm decisions}(t)}{ E_{\rm decisions}}\right], & \text{if deadlocked},\\[6pt]
K \, \left[\frac{N_{\rm pp}(t) + N_{\rm pp\%}(t)}{E_{\rm pp}} + \frac{N_{\rm decisions}(t)}{E_{\rm decisions}} + \frac{t}{E_{\rm seconds}} \right], & \text{otherwise}.
\end{cases}
\end{equation}

In practice, we found that $K=4000$ worked best for us. $E_{\rm pp}$, $E_{\rm decisions}$ and $E_{\rm seconds}$ are estimated based on the Cost Table Agent performances.

\subsection{DQN implementation}

In this research, we tested the vanilla DQN agent described in \cite{Mnih} and implemented some additional promising extensions to improve the agent's performance. This section briefly describes the implemented extensions. The exact hyperparameters used for the performance can be found in \ref{tab:ddqn_hyperparameters}. In \cite{Mnih}, the authors showed that the vanilla DQN-agent performed better in 43 of 49 games as the best previously known reinforcement learning baselines, setting a new standard in the industry. However, further investigations of the DQN showed that it is prone to be overly optimistic when estimating Q-values when using Q-learning. Hence, it overestimates the Q-values. This over-optimism could hinder the convergence to the agent's optimum, or the agent might not even converge at all. In \cite{HasseltGS15}, the authors showed that the over-optimism could be counteracted by applying Double Q-learning rather than Q-learning. The resulting agent is called DDQN and requires almost no changes to the existing implementation.

Moreover, in DQN, the baseline approach to handle the exploration-exploitation dilemma is the straightforward epsilon-greedy strategy. This strategy relies on random permutations of the sequence of actions to explore the environment. However, in some environments, where the reward is very sparse, it can be challenging to explore the environment sufficiently using this strategy. Hence, we also tried to drive exploration using noisy networks as described in \cite{APMOGM17} by inducing noise when estimating the Q-values. Moreover, in this research, we also tested agents using a dueling network architecture, splitting the data stream in the neural network into two separate data streams as described in \cite{WangFL15}. One data stream estimates the value of the state, while the other data stream estimates the advantage for picking an action in that state. Finally, the most significant extension that showed the most robust performance during all tests is the prioritized experience replay (PER) as described in \cite{schaul2015prioritized}. The PER samples the experiences prioritized based on the temporal difference error rather than sampling uniformly.

\begin{table}[t]
\begin{center}
    \begin{tabular}{|c c|} 
     \hline
     Parameter & Value \\ 
     \hline
     Replay memory size & 1e5 \\
     Batch size & 64 \\
     Gamma & 0.99 \\
     Learning rate & 0.001 \\
     Target update rate & 24 \\
     Update rate & 4 \\
     & \\
     Epsilon & 1.0 \\
     Epsilon decay & 0.9995 \\
     Epsilon min & 0.01 \\
     & \\
     NN nb layers & 2 \\
     FC1 units & 64 \\
     FC2 units & 32 \\
     \hline
    \end{tabular}
\end{center}
\caption{DDQN training hyperparameters.}
\label{tab:ddqn_hyperparameters}
\end{table}

\section{Experimental results}
\label{sec:results}

\begin{table}[ht]
\centering
\bgroup
\def\arraystretch{1.1} 
    
\begin{tabular}{|l|l||r|r|r|r|}
\hline
Environment & Agent & 1 AGV & 2 AGVs & 3 AGVs & 4 AGVs \\
\hline \texttt{Mayer-Classen-Endisch} & \texttt{FIFO}& & 50  \quad & &  \\ 
 & \texttt{Nearest Neighbor} & & 50  \quad & &  \\ 
 & \texttt{Cost Table} & & 50  \quad & &  \\ 
 & \texttt{DDQN} & & 0  \quad & &  \\ 
\hline \texttt{1-machine-big} & \texttt{FIFO}& & 83  \quad & &  \\ 
 & \texttt{Nearest Neighbor} & & 48  \quad & &  \\ 
 & \texttt{Cost Table} & & 11  \quad & &  \\ 
 & \texttt{DDQN} & & 0  \quad & &  \\ 
\hline \texttt{3-machines-loop} & \texttt{FIFO}& 153  \quad & 81  \quad & 56  \quad &  \\ 
 & \texttt{Nearest Neighbor} & 153  \quad & 80  \quad & 57  \quad &  \\ 
 & \texttt{Cost Table} & 120  \quad & 82  \quad & 55  \quad &  \\ 
 & \texttt{DDQN} & 0  \quad & 0  \quad & 0  \quad &  \\ 
\hline \texttt{2-machines-grid} & \texttt{FIFO}& 31  \quad & 25  \quad & 25  \quad & 25  \quad  \\ 
 & \texttt{Nearest Neighbor} & 31  \quad & 25  \quad & 24  \quad & 22  \quad  \\ 
 & \texttt{Cost Table} & 20  \quad & 25  \quad & 24  \quad & 22  \quad  \\ 
 & \texttt{DDQN} & 0  \quad & 0  \quad & 0  \quad & 0  \quad  \\ 
\hline \texttt{4-machines-grid} & \texttt{FIFO}& 41  \quad & 25  \quad & 25  \quad & 25  \quad  \\ 
 & \texttt{Nearest Neighbor} & 41  \quad & 25  \quad & 25  \quad & 25  \quad  \\ 
 & \texttt{Cost Table} & 37  \quad & 25  \quad & 25  \quad & 24  \quad  \\ 
 & \texttt{DDQN} & 0  \quad & 0  \quad & 0  \quad & 0  \quad  \\ 
\hline \texttt{6-machines-grid} & \texttt{FIFO}& 53  \quad & 29  \quad & 25  \quad & 25  \quad  \\ 
 & \texttt{Nearest Neighbor} & 53  \quad & 29  \quad & 25  \quad & 25  \quad  \\ 
 & \texttt{Cost Table} & 48  \quad & 28  \quad & 25  \quad & 25  \quad  \\ 
 & \texttt{DDQN} & 0  \quad & 0  \quad & 0  \quad & 0  \quad  \\ 
\hline
\end{tabular}

\egroup
\caption{Optimal source clock}
\label{tab:results_optimal_clock}
\end{table}
    
\begin{table}[ht]
\centering
\bgroup
\def\arraystretch{1.1} 
    
\begin{tabular}{|l|l||r|r|r|r|}
\hline
Environment & Agent & 1 AGV & 2 AGVs & 3 AGVs & 4 AGVs \\
\hline \texttt{Mayer-Classen-Endisch} & \texttt{FIFO}& & \textbf{71.8} \quad & &  \\ 
 & \texttt{Nearest Neighbor} & & \textbf{71.8} \quad & &  \\ 
 & \texttt{Cost Table} & & \textbf{71.8} \quad & &  \\ 
 & \texttt{DDQN} & & \textbf{71.8} \quad & &  \\ 
\hline \texttt{1-machine-big} & \texttt{FIFO}& & 23.6  \quad & &  \\ 
 & \texttt{Nearest Neighbor} & & 74.9  \quad & &  \\ 
 & \texttt{Cost Table} & & 143.3  \quad & &  \\ 
 & \texttt{DDQN} & & \textbf{143.4} \quad & &  \\ 
\hline \texttt{3-machines-loop} & \texttt{FIFO}& 23.4  \quad & 44.2  \quad & 63.8  \quad &  \\ 
 & \texttt{Nearest Neighbor} & 23.4  \quad & 44.8  \quad & 62.8  \quad &  \\ 
 & \texttt{Cost Table} & \textbf{26.9} \quad & 43.7  \quad & \textbf{65.0} \quad &  \\ 
 & \texttt{DDQN} & 26.3  \quad & \textbf{45.4} \quad & 60.2  \quad &  \\ 
\hline \texttt{2-machines-grid} & \texttt{FIFO}& 115.8  \quad & 143.7  \quad & \textbf{143.8} \quad & 143.7  \quad  \\ 
 & \texttt{Nearest Neighbor} & 115.8  \quad & \textbf{143.8} \quad & \textbf{143.8} \quad & 143.7  \quad  \\ 
 & \texttt{Cost Table} & \textbf{123.1} \quad & \textbf{143.8} \quad & \textbf{143.8} \quad & 143.7  \quad  \\ 
 & \texttt{DDQN} & 115.2  \quad & 139.2  \quad & \textbf{143.8} \quad & \textbf{143.8} \quad  \\ 
\hline \texttt{4-machines-grid} & \texttt{FIFO}& 87.4  \quad & \textbf{143.4} \quad & 143.4  \quad & \textbf{143.5} \quad  \\ 
 & \texttt{Nearest Neighbor} & 87.4  \quad & \textbf{143.4} \quad & \textbf{143.5} \quad & \textbf{143.5} \quad  \\ 
 & \texttt{Cost Table} & \textbf{95.7} \quad & \textbf{143.4} \quad & \textbf{143.5} \quad & 143.3  \quad  \\ 
 & \texttt{DDQN} & 88.8  \quad & 128.9  \quad & 138.9  \quad & 134.1  \quad  \\ 
\hline \texttt{6-machines-grid} & \texttt{FIFO}& 67.5  \quad & 123.5  \quad & 143.1  \quad & \textbf{143.2} \quad  \\ 
 & \texttt{Nearest Neighbor} & 67.5  \quad & 123.5  \quad & \textbf{143.2} \quad & \textbf{143.2} \quad  \\ 
 & \texttt{Cost Table} & \textbf{74.1} \quad & \textbf{127.8} \quad & \textbf{143.2} \quad & \textbf{143.2} \quad  \\ 
 & \texttt{DDQN} & 68.6  \quad & 79.0  \quad & 117.4  \quad & 124.0  \quad  \\ 
\hline
\end{tabular}

\egroup
\caption{Throughputs in parts per hour (pph)}
\label{tab:results_throughput}
\end{table}

\begin{table}[ht]
\centering
\bgroup
\def\arraystretch{1.1} 
    
\begin{tabular}{|l|l||r|r|r|r|}
\hline
Environment & Agent & 1 AGV & 2 AGVs & 3 AGVs & 4 AGVs \\
\hline \texttt{Mayer-Classen-Endisch} & \texttt{FIFO}& & 862  \quad & &  \\ 
 & \texttt{Nearest Neighbor} & & 862  \quad & &  \\ 
 & \texttt{Cost Table} & & 862  \quad & &  \\ 
 & \texttt{DDQN} & & \textbf{863} \quad & &  \\ 
\hline \texttt{1-machine-big} & \texttt{FIFO}& & 284  \quad & &  \\ 
 & \texttt{Nearest Neighbor} & & 899  \quad & &  \\ 
 & \texttt{Cost Table} & & 1720  \quad & &  \\ 
 & \texttt{DDQN} & & \textbf{1721} \quad & &  \\ 
\hline \texttt{3-machines-loop} & \texttt{FIFO}& 281  \quad & 531  \quad & 768  \quad &  \\ 
 & \texttt{Nearest Neighbor} & 281  \quad & 538  \quad & 755  \quad &  \\ 
 & \texttt{Cost Table} & \textbf{326} \quad & 525  \quad & \textbf{782} \quad &  \\ 
 & \texttt{DDQN} & 317  \quad & \textbf{547} \quad & 725  \quad &  \\ 
\hline \texttt{2-machines-grid} & \texttt{FIFO}& 1392  \quad & \textbf{1726} \quad & 1726  \quad & 1726  \quad  \\ 
 & \texttt{Nearest Neighbor} & 1392  \quad & \textbf{1726} \quad & 1726  \quad & \textbf{1727} \quad  \\ 
 & \texttt{Cost Table} & \textbf{1479} \quad & \textbf{1726} \quad & 1726  \quad & \textbf{1727} \quad  \\ 
 & \texttt{DDQN} & 1383  \quad & 1672  \quad & \textbf{1727} \quad & \textbf{1727} \quad  \\ 
\hline \texttt{4-machines-grid} & \texttt{FIFO}& 1051  \quad & \textbf{1724} \quad & \textbf{1724} \quad & 1724  \quad  \\ 
 & \texttt{Nearest Neighbor} & 1051  \quad & \textbf{1724} \quad & \textbf{1724} \quad & 1724  \quad  \\ 
 & \texttt{Cost Table} & \textbf{1151} \quad & \textbf{1724} \quad & \textbf{1724} \quad & \textbf{1725} \quad  \\ 
 & \texttt{DDQN} & 1068  \quad & 1551  \quad & 1670  \quad & 1611  \quad  \\ 
\hline \texttt{6-machines-grid} & \texttt{FIFO}& 812  \quad & 1485  \quad & \textbf{1722} \quad & \textbf{1723} \quad  \\ 
 & \texttt{Nearest Neighbor} & 812  \quad & 1485  \quad & \textbf{1722} \quad & \textbf{1723} \quad  \\ 
 & \texttt{Cost Table} & \textbf{892} \quad & \textbf{1537} \quad & \textbf{1722} \quad & \textbf{1723} \quad  \\ 
 & \texttt{DDQN} & 828  \quad & 951  \quad & 1413  \quad & 1491  \quad  \\ 
\hline
\end{tabular}

\egroup
\caption{Total parts produced over 12h}
\label{tab:results_total_parts}
\end{table}

\subsection{Per-scenario results}

The benchmark has been run on each scenario for 12 hours with 4 different agents: 3 static (FIFO, Nearest Neighbor, Cost Table) and 1 trained (DDQN). Before running the simulation, we determined the optimal source clock $C_{\rm source}*$ for each static agent. The main metric that we monitor is the overall throughput, defined in parts per hour (pph). We also measured the total amount of parts produced in 12 hours.

Optimal source clocks can be found in table \ref{tab:results_optimal_clock}, throughput results in table \ref{tab:results_throughput}, and total parts produced results in table \ref{tab:results_total_parts}.

\subsubsection{Mayer-Classen-Endisch}

\begin{figure}[ht]
  \includegraphics[width=1.0\textwidth]{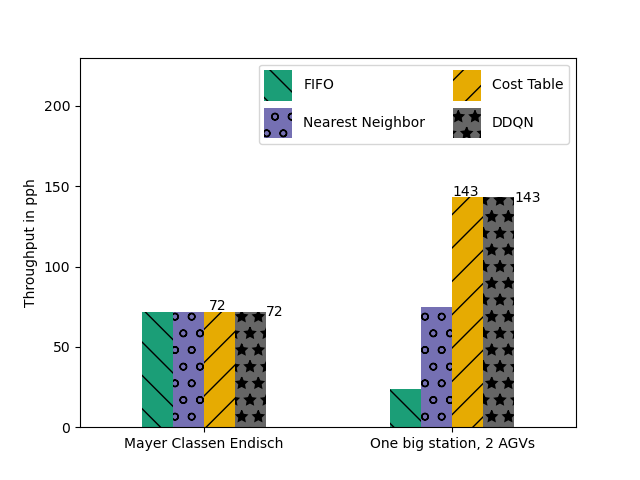}
  \caption{Results on Mayer-Classen-Endisch and $1$-machine-big environments.}
  \label{fig:simple_cases_results}
\end{figure}

The first environment we trained on has been the one described by \cite{Mayer2021}. Results are shown in figure \ref{fig:simple_cases_results}. The trained agent reached a throughput of 71.8 parts/hour over 12 hours (863 parts). The maximum throughput in this case is 72 parts/hours (864 parts). This environment has been a good first example, since we have been able to reproduce 2 results described by \cite{Mayer2021}:
\begin{itemize}
    \item The agent starts with a ramp-up process before reaching the optimal throughput. This explains the difference of 1 produced part observed over the 12 hours.
    \item The final policy learned by the agent is the same, namely, the AGV loops through the production plant in a certain order: $\text{Source} \rightarrow M1_{\rm input} \rightarrow M1_{\rm output} \rightarrow M2_{\rm input} \rightarrow M2_{\rm  output} \rightarrow \text{Sink} \rightarrow \text{Source} \rightarrow \ldots$
\end{itemize}

Finally, it is also worth mentioning that the baseline agents (FIFO, Nearest Neighbor and Cost Table) have also been able to achieve the same performance (862 parts produced). The main benefit of the DDQN agent is its robustness to the source clock.

\subsubsection{1-machine-big}

With this configuration, we expected the FIFO agent to be tricked, while a Cost Table approach was supposed to tackle the problem raised by the long distances in the production plant. We confirmed this intuition, since they reached 23 parts/hour and 143.4 parts/hour respectively, the optimal throughput in the equilibrium state being 144 parts/hour. The same performance has been achieved by the trained agent. Results can be compared in figure \ref{fig:simple_cases_results}.

For both Cost Table and DDQN approaches, the final policy in the equilibrium state is to have an AGV looping between the Source and the Machine entry and the other AGV looping between the Machine exit and the Sink, thus avoiding the long drive between Machine entry and exit.

It is also worth mentioning that the Nearest Neighbor approach gets intermediate results (74.9 parts/hour) which confirm that all static approaches are not equivalent even in small scenarios.

\subsubsection{3-machines-loop}

\begin{figure}[ht]
  \includegraphics[width=1.0\textwidth]{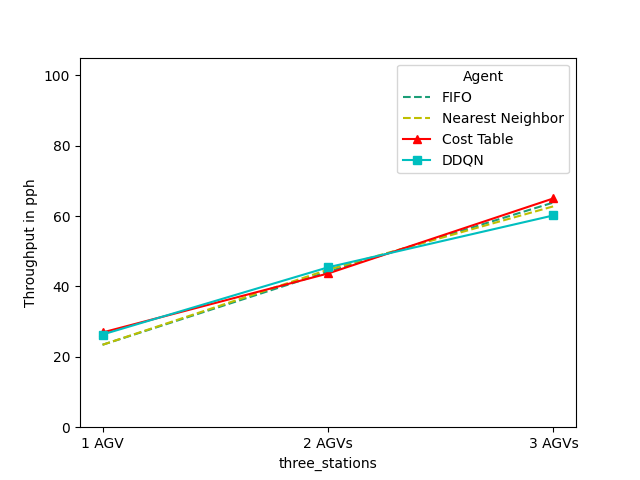}
  \caption{Results on the $3$-machines-loop environment.}
  \label{fig:3_machines_results}
\end{figure}

The difficulty in this scenario is to manage the AGV(s) in order not to overload the first machine. Results show that in the end, the different agents get on average similar results (see figure \ref{fig:3_machines_results}). In the details, Cost Table is slightly better than DDQN with 1 AGV (26.9 vs 26.3 pph), slightly worse with 2 AGVs (43.7 vs 45.4 pph) and significantly better with 3 AGVs (65.0 vs 60.2 pph). This shows the difficulties the DDQN agent has to manage multiple AGVs with our framework. We will discuss this aspect further in \ref{sssec:several_agvs}.

\subsubsection{6-machines-grid}

\paragraph{2-machines scenario}

\begin{figure}[ht]
  \includegraphics[width=1.0\textwidth]{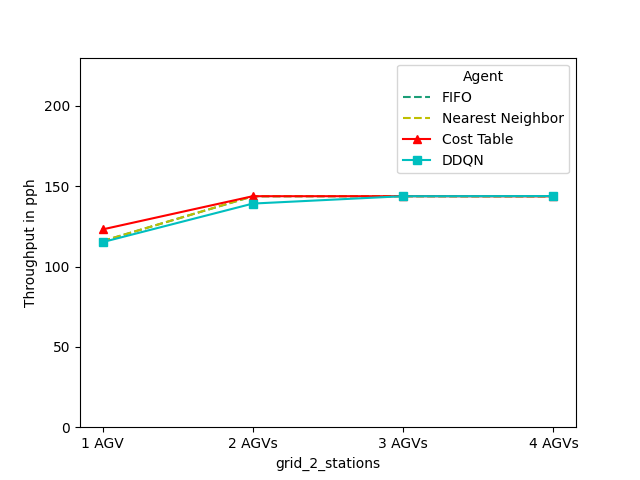}
  \caption{Results on the grid-$2$-machines environment.}
  \label{fig:grid_2_machines_results}
\end{figure}

The 2-machines scenario seems quite easy to tackle for the static agents (see figure \ref{fig:grid_2_machines_results}). The Cost Table approach performs better with 1 AGV but starting from 2 AGVs, the processing time is more limiting than transportation, which means that even a sub-optimal policy is able to obtain the best theoretical throughput (144 pph). 

The DDQN agent is able to reach FIFO/Nearest Neighbor performances with 1 AGV (115 pph) but is not optimal since Cost Table performs significantly better (123 pph). With 2 AGVs it is not optimal either, performing at 139 pph even though all static agents are already reaching 144 pph. Above that (3 and 4 AGVs), comparisons don't make sense, since all agents reach the limit.

\paragraph{4-machines scenario}

\begin{figure}[ht]
  \includegraphics[width=1.0\textwidth]{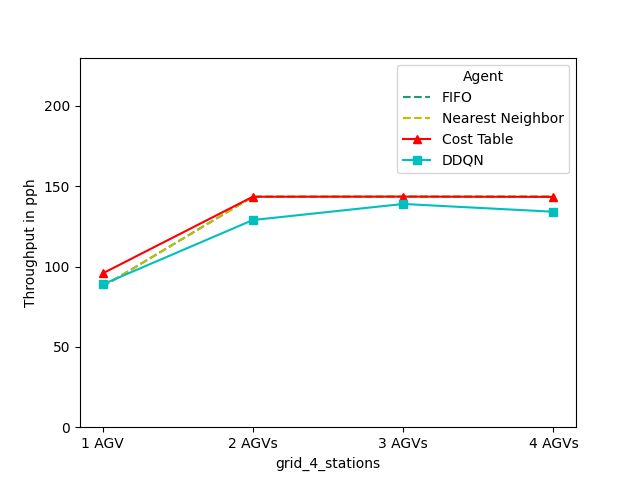}
  \caption{Results on the \texttt{grid-4-machines} environment.}
  \label{fig:grid_4_machines_results}
\end{figure}

The 4-machines scenario is similar to the previous one (see figure \ref{fig:grid_4_machines_results}). Cost Table performs better than other approaches with 1 AGV (96 pph vs 89) and then all static agents reach the limit (144 pph). 

However, this scenario starts to show the limitation of the DDQN agent. It is able to learn how to manage the AGVs to transport parts and still avoid deadlocks, but in a sub-optimal way. The best results are achieved with 3 AGVs (139 pph), a better performance than when a 4th AGV was available (134 pph). 

\paragraph{6-machines scenario}

\begin{figure}[ht]
  \includegraphics[width=1.0\textwidth]{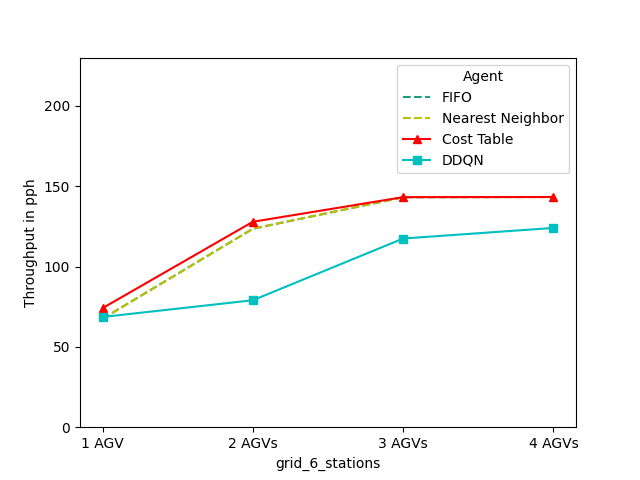}
  \caption{Results on the \texttt{grid-6-machines} environment.}
  \label{fig:grid_6_machines_results}
\end{figure}

In the 6 machines scenario, it is even more difficult for the DDQN to reach optimal performances -or at least similar to static agents- (see figure \ref{fig:grid_6_machines_results}). It is worth mentioning that even if the DDQN performs worse than Cost Table in all cases (69 vs 74 pph with 1 AGV, 79 vs 128 pph with 2, 117 vs 143 with 3 and 124 vs 143 with 4), it is still able to run the 12 hours without deadlocks.

\subsection{Improvement perspectives}

\subsubsection{Improve the observation state}

Our final observation state does not provide all the information needed by the agent to take the optimal decision. In particular, with our approach, the DDQN agent can't take advantage of the exact position of the AGVs and the transportation times. Our efforts to include this information in the observation state have not yet been successful. However, this information is available and used by the Cost Table agent, which can explain its better performances.

\subsubsection{Improve reward signal and deadlock punishment}

Another learning of our study is the sensitivity of the DDQN agent to the reward signal, especially to the deadlock punishment signal. In the end, our agent learned how to avoid running into a deadlock but at the cost of been too protective in some cases (i.e., it is more risky and not enough rewarded to start processing a new part as compared to wait for the production plant to be less full). This balance has been tough to configure and is still not entirely satisfying.

\subsubsection{Improve management of several AGVs}
\label{sssec:several_agvs}

The study has shown the limitation of our approach when it comes to managing several AGVs at the same time. We believe that there is room for improvement in the communication protocol between the Agent and the AGVs. Our intuition is that the current set of actions the Agent can take is too limited. We had difficulties into making the Agent choose both the Station and the AGV that must perform the action. 

\section{Conclusions and Outlook}
\label{sec:conclusions}
In this paper, we investigated the deployment of Deep-Q based DRL agents to job-shop scheduling problems in modular production facilities using discrete event simulations for the environment. Here are the key findings of our study:
\begin{itemize}
\item We have developed a simulation that runs fast enough to train DRL agents. We have established a framework to configure environments of different complexity, on which we can additionally run a baseline of three static agents.

\item For each environment, we trained a DDQN agent that is able to run the production plant deadlock-free. The trained agent has learned some complex strategies (beating FIFO in $1$-machine-big and handling a loop in $3$-machines-loop environment). 

\item We proved that it is possible to apply DRL to the vehicle management problem. From this, there is room for improvement, especially for managing complex scenarios with more AGVs and stations.

\item Beyond the scope of this paper, there exist various avenues for further exploration. For instance, we hypothesize that a DRL approach increases stability in the system. This is suggested by the fact the DDQN agent is robust against the source clock, while static agents need additional precautions. We also performed simulations in which we introduced noise (in the source clock, the transports, the parts transfers and the processing). The trained agents tend to be more robust in those cases, avoiding deadlocks.
\end{itemize}

With these findings, we have laid out a foundation for future studies that can build upon these findings. Both, the agents, and the environment, can be improved to facilitate environments of a richer variety, and particular focus can be dedicated to further investigating the robustness of the DRL agents in comparison to the heuristic approaches. Also, the issue of dead-lock avoidance could be investigated more thoroughly. As for the environment and apart from increasing the complexity by means of adding stations, a more diverse work-flow, or more AGVs, one could furthermore envision an environment in which the AGVs themselves are controlled by DRL agents, considering additional factors such as battery status, nearest charging station, position of the other AGVs, etc. However, all these avenues are left unaddressed in this study, as the purpose of this paper is to establish a first, preliminary exploration of the feasibility of Deep-Q based DRL approaches in modular production environments.

\section{Acknowledgements and references}
\label{sec:acknowledgements}
Partial funding through the Austrian Research Promotion Agency FFG, Kleinprojekt Nr. 883243, is kindly acknowledged.
Parts of the findings of this article are also given in \cite{Hasenbichler2021}.

\bibliography{literature}
\end{document}